# Air Quality Measurement Based on Double-Channel Convolutional Neural Network Ensemble Learning

Zhenyu Wang, Wei Zheng, Chunfeng Song

School of Control and Computer Engineering, North China Electric Power University, Beijing, China

{zywang, wzheng, scf}@ncepu.edu.cn

## Abstract

Environmental air quality affects people's life, obtaining real-time and accurate environmental air quality has a profound guiding significance for the development of social activities. At present, environmental air quality measurement mainly adopts the method that setting air quality detector at specific monitoring points in cities and timing sampling analysis, which is easy to be restricted by time and space factors. Some air quality measurement algorithms related to deep learning mostly adopt a single convolutional neural network to train the whole image, which will ignore the difference of different parts of the image. In this paper, we propose a method for air quality measurement based on double-channel convolutional neural network ensemble learning to solve the problem of feature extraction for different parts of environmental images. Our method mainly includes two aspects: ensemble learning of double-channel convolutional neural network and self-learning weighted feature fusion. We constructed a double-channel convolutional neural network, used each channel to train different parts of the environment images for feature extraction. We propose a feature weight self-learning method, which weights and concatenates the extracted feature vectors, and uses the fused feature vectors to measure air quality. Our method can be applied to the two tasks of air quality grade measurement and air quality index (AQI) measurement. Moreover, we build an environmental image dataset of random time and location condition. The experiments show that our method can achieve nearly 82% accuracy and a small mean absolute error (MAE) on our test dataset. At the same time, through comparative experiment, we proved that our proposed method gained considerable improvement in performance compared with single channel convolutional neural network air quality measurements.

## 1 Introduction

Environmental air quality is closely related to human production and life. The decline of air quality is likely to cause ecological damage and induce human diseases. At present, air quality monitoring mainly adopts the method of setting up monitoring stations in several specific locations in the city, using the air quality detector to regularly sample and measure air pollutants, and finally obtaining the air quality index through calculation and analysis. This method is easy to be limited by time and space, can only obtain air quality at specific monitoring points in specific time. It is difficult to obtain the air quality information of the random location in real time, and the measurement cost is high. How to obtain the air quality index in real time and accurately is a subject worth studying.

Image-based air quality measurement is a method that use image processing algorithm to extract environmental image features and estimate air quality index based on image features. In recent years, with the rapid development of deep learning technology, using deep learning technology to complete identification, detection and other tasks is efficient. Environmental images under different air quality grades are often different to some extent, therefore, it is feasible and valuable to use deep convolutional neural network to extract features of environmental images and complete the measurement of real-time air quality index at random site. Compared with the traditional air quality measurement method, the air quality measurement based on image and deep learning can obtain the air quality at any time and any place, which has the advantages of real-time and low cost, has been widely concerned by the academic circle in recent years.

At present, the existing air quality measurement methods related to image or deep learning are mainly divided into two types: method based on traditional image processing or deep learning. The methods which based on traditional image processing [1,7], are use traditional machine learning algorithms for feature extraction, such as edge detection, direction gradient histogram, etc. The extracted features are analyzed and calculated to get air quality measurement values. The image-based deep learning methods [2,3,4,5] generally train the deep convolutional neural network model, extracts the environmental image features, and calculates the air quality measurement values.

Most of the above air quality measurement algorithms based on image and deep learning adopt convolutional neural network to extract features of the whole image. However, due to the complexity of the environmental image content composition, the change law of air pollution between the sky part and the building part are different. Under the condition of using the same convolutional neural network for feature extraction, the differences are often ignored. Considering the particularity of the problem, we proposed a double-channel convolutional neural network to extract different parts of the image features separately and measure the air quality index shown in the environmental image. Compared with other methods, our method is more targeted to extract different parts of environmental image features, focuses more on the selection of excellent features.

This paper proposes an air quality measurement algorithm based on double-channel convolutional neural network ensemble learning, we first construct a kind of suitable double-channel convolutional neural network structure for air quality measurement. Start from the idea of ensemble learning, we use two channel convolutional neural networks respectively extract the feature in the different part of environment image. Secondly, considering that different parts of the image have different effect weight on the final recognition result, we propose a weighted feature fusion method to fuse the feature vectors extracted from the two channels. Finally, the integrated global feature vector is used to measure the air quality of the environment image. In addition, we also propose a self-learning method of weight to find the appropriate feature fusion weight. In terms of training and testing, we constructed an environmental image dataset with different locations, times and air qualities by means of manual shooting and collection. Through testing experiment, our method can achieve a certain accuracy and a small MAE in our dataset. The algorithm diagram is shown in figure 1.

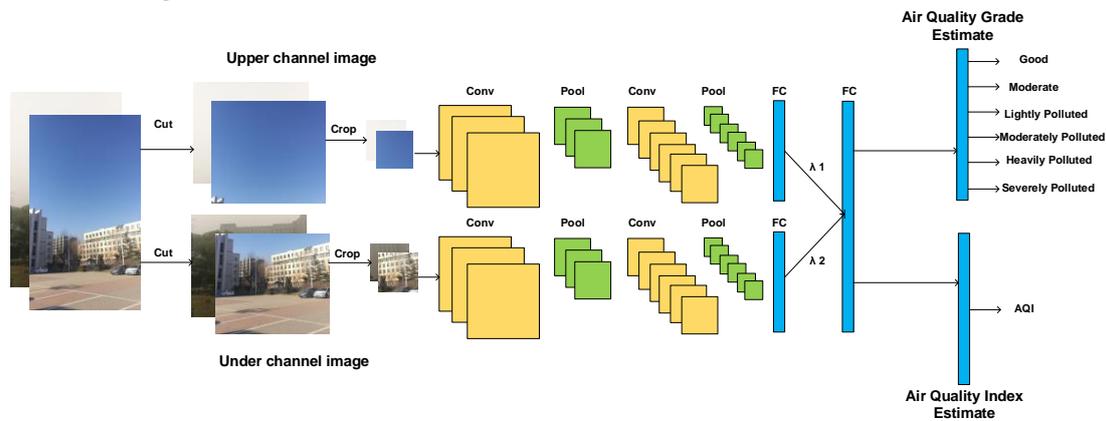

Figure 1. Air quality measurement based on double-channel convolutional neural network ensemble learning algorithm diagram

The main contributions of this paper:
1. We construct a double-channel convolutional neural network structure to perform feature extraction for different parts of the environmental image.
2. We propose a weighted feature fusion method, and a feature weight self-learning method to select excellent features.
3. We apply the double-channel convolutional neural network and weighted feature fusion to the classification and regression tasks, complete the tasks of air quality grade measurement and air quality index measurement.
4. Through experiments, we prove the effectiveness of the proposed method, and demonstrate the influence of different weights and different network structure on system performance.

The rest of this paper is organized as follows. Section 2 is related work, mainly introduces the development of deep learning and related research on air quality measurement. Section 3 introduces the air quality measurement algorithm based on double-channel convolutional neural network ensemble learning, mainly introduces the structure of the double-channel convolutional neural network, the weighted feature fusion and the self-learning method of feature weight. Section 4 is the experimental part, mainly introduces our training and testing methods, shows our experimental results, and compares our system with different network structure, different feature weights ratio, analyzes the existing problems. Section 5 is the conclusion, which concludes the paper and describes the main direction of the future work.

## 2　Related Work

The research on deep learning can be traced back to 1989, when LeCun applied BP algorithm to multi-layer neural network [8]. With the LeNet-5 model proposed by LeCun in 1998 [9], the basic structure of deep neural network was formed. In 2006, Geoffrey Hinton, a professor at the university of Toronto, formally proposed the concept of deep learning [10], and deep learning has entered a period of rapid development. Alex proposed AlexNet in 2012 [11], built the first big convolutional neural network, and adopted the ReLu activation function instead of Sigmod, managed to avoid the problem of gradient disappeared in neural network training, the performance of image recognition is much better than traditional methods. VGG [12], GoogleNet [13], ResNet [14] and other network structures were proposed one after another, which further enhanced the feature extraction ability of deep convolutional neural network. Using deep convolutional neural network to extract image features and complete classification and recognition has become the primary choice and important research direction for more and more researchers. In essence, air quality measurement is an image-based classification or regression task. Deep convolutional neural network can be used to effectively extract environmental image features and complete the identification task of air quality as shown in the image. In recent years, the measurement of air quality using deep learning method has attracted much attention in academic circles.

In the study of air quality measurement related to deep learning, Chao Zhang [2] built a convolutional neural network, improved the convolutional layer activation function and classification layer activation function, proposed an improved activation function of convolutional neural network EPAPL, and used a Negative Log-Log Ordinal Classifier to replace softmax Classifier in the classification layer, used the environment image to train their network model for classification prediction, completed measurement task of $PM_{2.5}$ and $PM_{10}$ in six grades; Avijoy Chakma et al [3] used convolutional neural network training images for feature extraction, combined with random forest classification, and classified the air quality shown in the images into three grades of good, middle, and bad. Nabin Rijal [4] adopts a method of neural network ensemble learning, they used three different convolutional neural networks, VGG16, InceptionV3 and Resnet50 to respectively conduct regression training on image $PM_{2.5}$ values. The predicted values of the input $PM_{2.5}$ of the three networks were input as feature into a feedforward network for training to predict the image $PM_{2.5}$ values. Jian Ma [5] combined the dark channel prior theory [6], firstly extracted the dark channel from the image, trained two convolutional neural networks respectively with the original image and the dark channel images, and identified the good and bad air quality of the image in three grades. Xiaoguang Chen et al [7] proposes a traditional image processing algorithms and deep learning combining approach. First, they counted distribution characteristics of image pixel values, statistics the proportion of high brightness points (pixel value > 128) to all pixel points of each image, use edge detector statistics all image the proportion of edge points to all pixels. The two values of each image as input feature to train the BP neural network, to predict air quality index value.

Considering the different composition information of different parts of the environmental image, we constructed a double-channel convolutional neural network based on the method of deep convolutional neural network and the idea of ensemble learning to extract features from different parts of the image, proposed a double-channel convolutional neural network ensemble learning algorithm for air quality measurement.

## 3 Double-Channel Convolutional Neural Network Ensemble Learning Algorithm for Air Quality Measurement

Alex proposed AlexNet [11] has achieved good performance in image recognition tasks, in view of the air quality measurement in essence for image recognition tasks, therefore, on the basis of AlexNet [11], we constructed a double-channel convolutional neural network for feature extraction of the sky and building parts of the environmental image, weighted and fused the extracted features, proposed a double-channel convolutional neural network ensemble learning algorithm for air quality measurement. It is composed of two feature extraction convolutional neural networks, a weighted feature fusion layer and a classification layer, as shown in table 1.

Table1. Double-channel convolutional neural network structure

| Upper channel image 64*64*3 | Under channel image 64*64*3 |
|---|---|
| Conv1a 5*5 128 stride=2 | Conv1b 5*5 128 stride=2 |
| Conv2a 5*5 128 stride=1 | Conv2b 5*5 128 stride=1 |
| Max_pool2a 3*3 stride=2 | Max_pool2b 3*3 stride=2 |
| Conv3a 5*5 128 stride=1 | Conv3b 5*5 128 stride=1 |
| Max_pool3a 3*3 stride=2 | Max_pool3b 3*3 stride=2 |
| Conv4a 3*3 192 stride=1 | Conv4b 3*3 192 stride=1 |
| Conv5a 3*3 192 stride=2 | Conv5b 3*3 192 stride=2 |
| FC6a 512 | FC6b 512 |
| FC6a*$\lambda_1$+FC6b*$\lambda_2$ ||
| FC7 1024 ||
| FC7 6 ||

### 3.1 Double-Channel Convolutional Neural Network Structure

The structure of the double-channel convolutional neural network is shown in table 1. It is composed of upper and under channel sub-convolutional neural networks, each channel convolutional neural network contains five convolution layers, two pooling layers and one fully connection layer. The first three convolution layers adopt 5*5 convolution kernel, and the last two convolution layers adopt 3*3 convolution kernel for feature extraction of image; Maximum pooling is used in each pooling layer to extract important features from down sampling; The 512-nodes fully connection layer is used to output feature vectors extracted from each network for feature fusion and prediction.

For different components of the environment image, the double-channel convolutional neural network adopts the strategy of ensemble learning to receive different parts of the image simultaneously in the upper and under channels for training. Before inputting the environment image into the double-channel convolutional neural network for training, the environment image should be preprocessed first, and the partial sky image and the partial building image should be segmented. For each image, the horizontal central axis average segmentation method is adopted to divide the image into the image mainly containing the upper half of the sky and the image mainly containing the under half of the building. Among them, the upper channel convolutional neural network focuses on feature extraction of the sky. In each round of iterative training, the images of the upper half with more sky elements after cutting are input into the upper channel convolutional neural network for training; The convolutional neural network of the under channel focuses on feature extraction of the building part. In each round of iterative training, the image of the under half with more building elements after cutting is input into the convolutional neural network of the under channel for training. After feature extraction at the fully connection layer of the last layer of each subnetwork, the feature vectors of the upper and under parts were weighted and fused, and the feature vectors containing complete features of the upper and under channels were used for recognition.

### 3.2 Weighted Feature Fusion and Weights Self-Learning

We found in our observation that, considering the images input of two channels, the content of some images of the sky is relatively simple, generally simple sky, or containing a small number of clouds or trees, the image complexity is relatively low; The content of the building is relatively rich in composition, and there are a variety of different buildings, roads and trees, etc., and the image complexity is relatively high. Due to the different complexity of the two images, the feature complexity extracted by the two channels is also different, and the effect weight on the final measurement result is different.

Therefore, considering that the image features of the upper and under channels may have different proportion of influence on the measurement results, we propose a method of weighted feature fusion. Before sending the output features of the two feature layers into the classification layer, the weighted feature fusion is carried out first. The weight value is multiplied by the output feature vectors of the upper and under channels by two constants, and then the two vectors are concatenated. The formula of feature fusion is as equation (1):

$$feat = \lambda_1 * feat_a + \lambda_2 * feat_b \qquad (1)$$

Where $\lambda_1$ and $\lambda_2$ are weight values of upper and under channels respectively; $feat_a$ and $feat_b$ are feature vectors extracted by upper and under channels respectively; $feat$ is global features after feature fusion.

On the basis of artificially assigning feature weights, we propose a self-learning method for feature weights, which also participate in training. In the initial stage, two weights $\lambda_1$ and $\lambda_2$ are set to 0.5 by adopting the strategy of balancing weights. In the training of the double-channel convolutional neural network, we only train other network parameters of the double-channel convolutional neural network with the weights value is frozen. After the training of the double-channel convolutional neural network, other parameters of the network should be frozen and the two weights are trained to find the appropriate feature ratio. In feature weights training, considering the proportional relationship between the two weights, we propose a weight loss constraint function to limit the training of weight values. The weight loss constraint function is defined as equation (2):

$$Loss_w = [1 - (\lambda_1 + \lambda_2)]^2 \tag{2}$$

When training the feature weight, the objective loss and the weight loss constraint function are combined to form the joint loss function, optimizing joint loss function to adjust the weights parameter value. Finally, multiply the two weights obtained after training a by the feature vectors extracted from each channel, and concatenate two weighted features.

### 3.3 Air Quality Measurement

For air quality measurement tasks, we start from the two directions of classification and regression, consider applying our double-channel convolutional neural network in two aspects, air quality grade measurement and air quality index measurement.

#### 3.3.1 Air Quality Grade Measurement

Air quality grade measurement is essentially a classification and recognition task. According to the 6 grades of air quality, the corresponding environmental images are divided into 6 categories and classified in the fully connection layer. Softmax was used for the activation function to conduct one-hot operation on all kinds of labels to obtain the predicted probability value of each grade, and the maximum probability was taken as the measurement result of each grade. At the same time, we put forward a calculation method of air quality index according to the prediction probability of each grade, like equation (3):

$$\text{AQI} = AQI_L + (AQI_H - AQI_L)(1 - P) \tag{3}$$

Where $AQI_H$ and $AQI_L$ are respectively the upper and lower limits of the predicted grade air quality index, and $P$ is the predicted probability of the predicted grade. According to the calculation, we can get the calculated value of air quality index.

#### 3.3.2 Air Quality Index Measurement

Based on the idea of regression, we consider the direct measurement of air quality index. Therefore, we add a 1-node fully connection layer after the above double-channel convolutional neural network, the AQI value corresponding to the environment image was used as the training label to conduct regression training. With directly measure the air quality index, we can calculate the air quality grade according to the air quality index value. The loss function adopts the mean square error between the predicted value and the labeled value:

$$Loss_R = \frac{1}{M}\sum_{i=1}^{M}(y_i - f(x_i))^2 \tag{4}$$

Where $M$ is the number of training images, $y_i$ is the labeled value of the ith image air quality index, and $f(x_i)$ is the predicted value.

# 4 Experiment

## 4.1 Dataset

In order to establish an effective dataset, we used the method of manual collection, to randomly shoot environmental images at different time of each day in the Beijing area, established a environmental image dataset. Images content distribution are basically the same, including the building and the sky two parts. At the same time, according to [17], we collected air quality index value of the current period for each image, and according to [16], we divided each image into corresponding grades, as the image label. The corresponding grade and the air quality index relations as shown in table 2. In this way, we built a dataset containing about 2500 environmental images under various air quality conditions.

Table2. Air quality grade table

| AQI | 0-50 | 51-100 | 101-150 | 151-200 | 201-300 | >300 |
|---|---|---|---|---|---|---|
| Grade | 1 | 2 | 3 | 4 | 5 | 6 |

In the study of this task, we screened the dataset. Due to the quality problem of the image itself, we manually removed the images with poor image quality, poor weather conditions, and inappropriate shooting time, such as sunset time light is poor, night by the influence of street lights. Considering the unbalanced dataset samples, images with low air quality index is much more than images with high air quality index, easy to cause an effect to training, makes the final model can be the index of high image recognition rate is too low, so we chose a relatively evenly number of each grades, finally formed a with 567 images dataset, partial sample image is shown in figure 2. Among them, 465 training images and 102 test images were included. The number of images of each category is shown in table 3, and the distribution of AQI is shown in figure 3.

Table3. Dataset air quality grade distribution

| Grade | 1 | 2 | 3 | 4 | 5 | 6 | ALL |
|---|---|---|---|---|---|---|---|
| **Train Set** | 98 | 94 | 70 | 90 | 79 | 34 | 465 |
| **Test Set** | 28 | 13 | 18 | 15 | 23 | 5 | 102 |
| **Complete Set** | 126 | 107 | 88 | 105 | 102 | 39 | 567 |

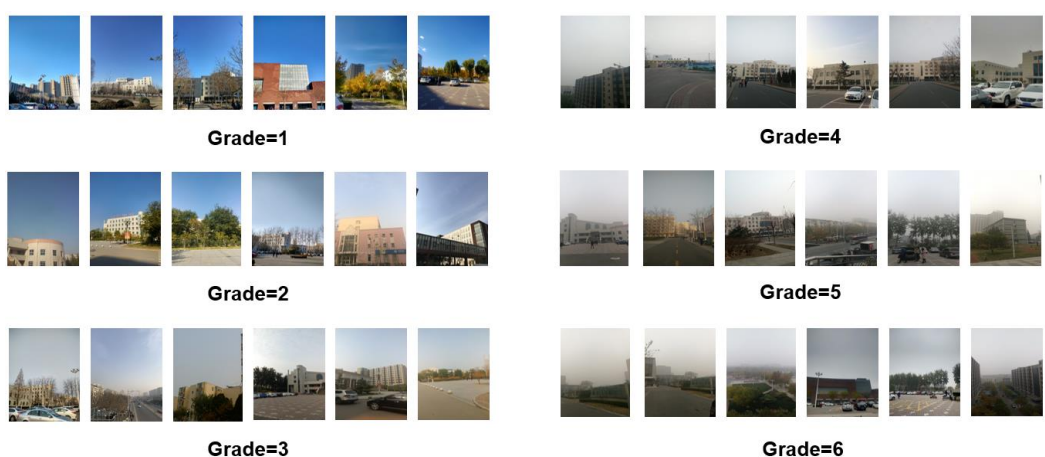

Figure2. Dataset images

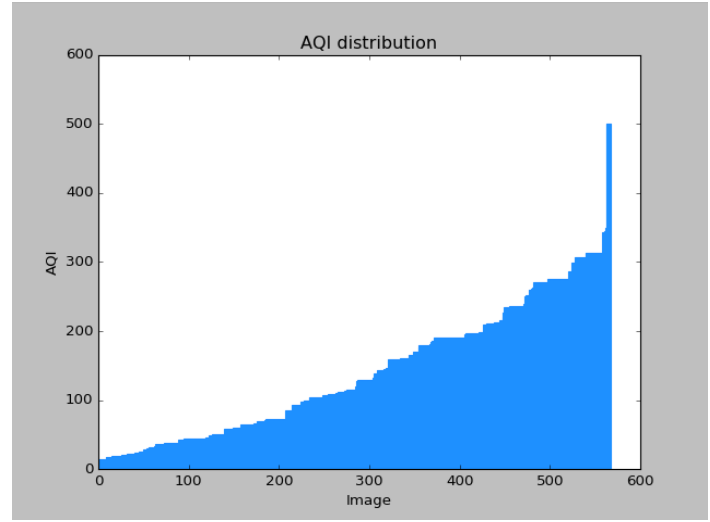

Figure3. Dataset air quality index distribution

### 4.2 Training

**4.2.1 Image Preprocessing**

Before training, we first resize each image and scale all images to the size of the short side 500 pixels and the long side to scale. In each iteration of training, we select a batch-size images, for each image, cut evenly based on horizontal central axis, and each image is divided into upper and under parts. Secondly, considering insufficient training data, we use random crop and random flip as data augmentation methods. Random crop is performed on two cropped images, and a 64*64 image block is cropped respectively. At the same time, the flip probability of 0.5 was used to carry out random horizontal transformation on the two images at the same time. Finally, for each batch-size, we obtained 64*64 images of batch-size*2 as the training data of this iteration, and the two cropped images on the same image were taken as a group, which were respectively fed into the upper and under channel convolutional neural network.

**4.2.2 Double-Channel Convolutional Neural Network Training**

At the end of image preprocessing, the obtained training data are simultaneously fed into the upper and under channel sub-convolutional neural network for training. The upper part containing more sky elements is fed into the upper channel convolutional neural network. The under part containing more building elements is fed into the under channel convolutional neural network. The loss function is calculated at the output layer. For the grade measurement problem, the loss function is the mean value of the cross entropy between the output value and the label value. For the AQI measurement problem, the loss function is the mean square error. The fusion weights $\lambda_1$ and $\lambda_2$ of feature fusion layer are frozen, and the negative feedback stochastic gradient descent is adopted to train other network parameters of the double-channel neural network. At the same time, we adopted dropout [15] with probability of 0.5 to prevent network overfitting at the last convolution layer of each channel convolutional neural network.

### 4.2.3 Feature Fusion Weight Training

When the training of the double-channel convolutional neural network meets the requirements and the loss function value no longer decreases significantly, we stopped the first step of training and froze the network parameters. Next, only two fusion weights of feature fusion layer were trained, and negative feedback stochastic gradient descent was used to update $\lambda_1$ and $\lambda_2$. After a certain number of iterations, the training was completed.

Training environment configuration, we adopt Intel Xeon e5-2650 v3@2.30GHz CPU, NVIDIA Tesla K40c GPU hardware environment; In terms of software, we adopt TensorFlow1.10.0 deep learning framework and Python3.5 programming language. In terms of training Settings, batch-size was 128, learning rate was 1-e4, the training period was about 3300 epochs, and the number of iterations was 11,000. The first 10,000 times are used to train the parameters of the double-channel convolutional neural network. The last 1000 iterations are used to adjust the feature weights.

## 4.3 Testing
### 4.3.1 Test Methods and Evaluation Criteria

We used test-time augmentation during the testing. That is, random crop was also used in the test, and voting mechanism and average mechanism were introduced. In the test, the above image preprocessing is carried out on the test image first, and each image is subject to 20 times of random crop without random horizontal transformation. For each image, we get 20 groups test data to feed into the trained model for prediction. Finally, 20 predicted results were obtained for each image. For the task of grades classification, the voting mechanism was adopted to take most of the predicted class grades in the 20 predicted results as the final prediction classification. For the AQI measurement task, the average mechanism was adopted, and the mean value of 20 predicted values was taken as the measurement result.

Two evaluation criteria, mean accuracy and mean absolute error (MAE), were used to evaluate the classification accuracy. The mean accuracy is shown in the following formula (5), that is, the ratio of the predicted correct sample number to the total sample number. where $N$ is the total number of test samples, $y_i^p$ is the ith sample predicted grade, $y_i^t$ is the ith sample labeled grade.

$$\text{Accuracy } = \frac{1}{N}\sum_{i=1}^{N} y_i^p = y_i^t \tag{5}$$

At the same time, because of the particularity of problem, labeling information collected from the nearest location of stations as well as the time is the most closed to the hour of the air quality index value, it is difficult to obtain the location accurate air quality index value, so the annotation information itself has a little error; In addition to the limitation of time and space, the measurement error of the measuring instrument itself makes the images of different grades at the critical point of air quality grade also have the problem of inaccuracy, and the difference is slight. Therefore, we use the mean absolute error (MAE) as the second evaluation standard, that is, to calculate the mean value of the absolute value of the difference between the predicted grade and the true grade of each

sample. The formula is as equation (6):

$$\text{MAE} = \frac{1}{N}\sum_{i=1}^{N}|y_i^p - y_i^t| \qquad (6)$$

For the air quality index regression problem, MAE is also used as the evaluation criteria to measure the mean deviation between the measured value and the real AQI value. At the same time, we introduce the mean deviation rate (MDR) as another evaluation criteria of index. The formula of MDR is as equation (7). Where $v_i^p$ is the ith a sample air quality index predicted value, $v_i^t$ is the ith a sample air quality index of true value.

$$\text{MDR} = \frac{1}{N}\sum_{i=1}^{N}\frac{|v_i^p - v_i^t|}{v_i^t} \qquad (7)$$

### 4.3.2 Double-Channel Convolutional Neural Network Air Quality Grade Recognition Performance Analysis

Based on the proposed double-channel convolutional neural network, we first tested model performance on recognition air quality grade (double channel equal weight for classify, DCEW-C) under the condition of equal weight, at the same time, we use the following methods for performance comparison: (1) single channel convolutional neural network training and testing on the whole image (baseline); (2) only the upper channel convolutional neural network is used for training and testing the upper part of the image; (3) only the under channel convolutional neural network is used for training and testing the under part of the image.

The test results of the above methods are shown in table 4. Due to the large similarity between adjacent air quality grade images, we also introduced the Neighbor accuracy as another reference evaluation criteria for grades recognition.

Table4. Comparison of air quality grade measurement performance of four convolutional neural network structures

| Method | Accuracy | Neighbor accuracy | MAE |
|---|---|---|---|
| Single Channel（Baseline） | 67.25% | 92.45% | 0.4245 |
| Single Upper Channel | 69.31% | 91.37% | 0.4039 |
| Single Under Channel | 74.80% | 89.80% | 0.4108 |
| DCEW-C（Our Method） | **80.88%** | **95.58%** | **0.2353** |

It can be seen from table 4 that the method using double-channel convolutional neural network is much better than the method only applied a single convolutional neural network in terms of accuracy, neighbor accuracy and MAE. For the single-channel convolutional neural network, the under channel convolutional neural network has achieved a high accuracy, but due to the complexity of the under half of the image, its neighbor accuracy is poor. For the single-channel convolutional neural network, due to the extraction of the whole image information, it performs better in the neighbor accuracy. The image features of the upper channel convolutional neural network are relatively simple, so it obtained the lower MAE.

For the proposed double-channel convolutional neural network (DCEW-C), because it takes into account different parts of the image information, and adopts the strategy of separately extracting ensemble learning, it has achieved a great performance improvement compared with the single-channel convolutional neural network. Compared with the optimal performance of each single channel convolutional neural network, the accuracy was improved by more than 6 percentage points, the neighbor accuracy was improved by more than 3 percentage points, and the MAE was reduced by 0.1686.

### 4.3.3 Double-Channel Convolutional Neural Networks with Different Feature Fusion Weights Performance Analysis

On the basis of the equal weight double-channel convolutional neural network for air quality grade recognition, we explore the performance under the weighted fusion of two-channel features. Considering from two aspects, we have conducted experiments with different assigned weights and with double-channel self-learning weight for classify (DCSLW-C). For the weight assignment, we studied the system performance with the feature weight ratio of upper and under channels at 3:7, 4:6, 5:5, 6:4 and 7:3 respectively. The experimental results are shown in table 5.

Table5. Comparison of air quality grades measurement performance of five kinds weights

| Ratio（Upper : Under） | Accuracy | Neighbor accuracy | MAE |
|---|---|---|---|
| 3:7 | 79.90% | **97.16%** | 0.2363 |
| 4:6 | 77.84% | 93.53% | 0.2990 |
| 5:5(DCEW-C) | 80.88% | 95.98% | 0.2353 |
| 6:4 | 78.43% | 92.05% | 0.3049 |
| 7:3 | 80.29% | 94.80% | 0.2539 |
| DCSLW-C | **81.86%** | 96.17% | **0.2225** |

As can be seen from table 5, system performance increases and decreases for different feature weights. For the improper feature weights, the performance of the system is obviously decreased compared with DCEW-C. Relatively speaking, some performance criteria have been improved with more appropriate feature weights. Therefore, adopting the method of weight self-learning is beneficial for the system to automatically find the appropriate feature weight. The double-channel convolutional neural network using the weight self-learning method (DCSLW-C) has improved its performance in terms of accuracy and MAE, compared with DCEW-C and the assigned weights feature fusion method.

### 4.3.4 Double-Channel Convolutional Neural Network Air Quality Index Measurement Performance Analysis

In addition, from the perspective of regression, we conducted experiments on the direct measurement of the air quality index (AQI) with double-channel convolutional neural network, and classified the images to corresponding grade according to the predicted results. Similarly, we respectively used double-channel self-learning weight for regression (DCSLW-R)、double-channel equal weight for regression (DCEW-R)、single channel for regression (SC-R) for AQI measurement experiments. The experimental results of AQI prediction are shown in figure 4. For the task of AQI measurement, the performance difference between the double-channel convolutional neural network with equal weight and self-learning weight is small, and they have respective advantages in MAE and MDR. It can be seen from the figure 4 that in the AQI measurement, the prediction of lower air quality index is more accurate, and the accuracy decreases with the increase of the index. Compared with single channel convolutional neural network, the measurement error of double channel convolutional neural network is obviously lower. The partial AQI prediction results of DCSLW-R are shown in figure 5.

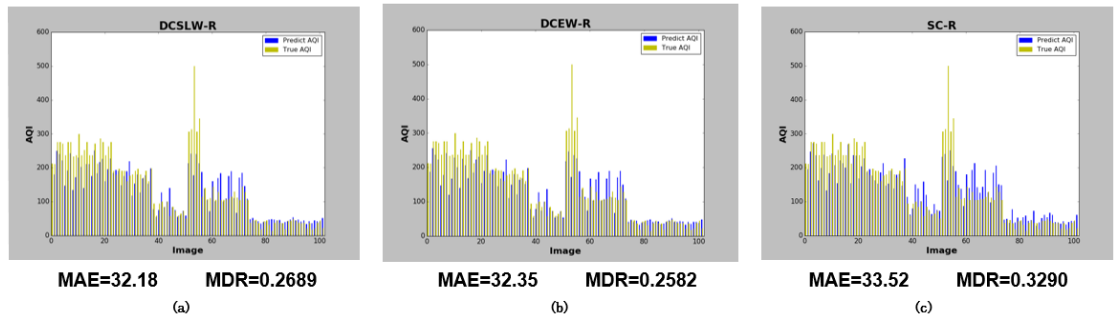

Figure4. AQI measurements result

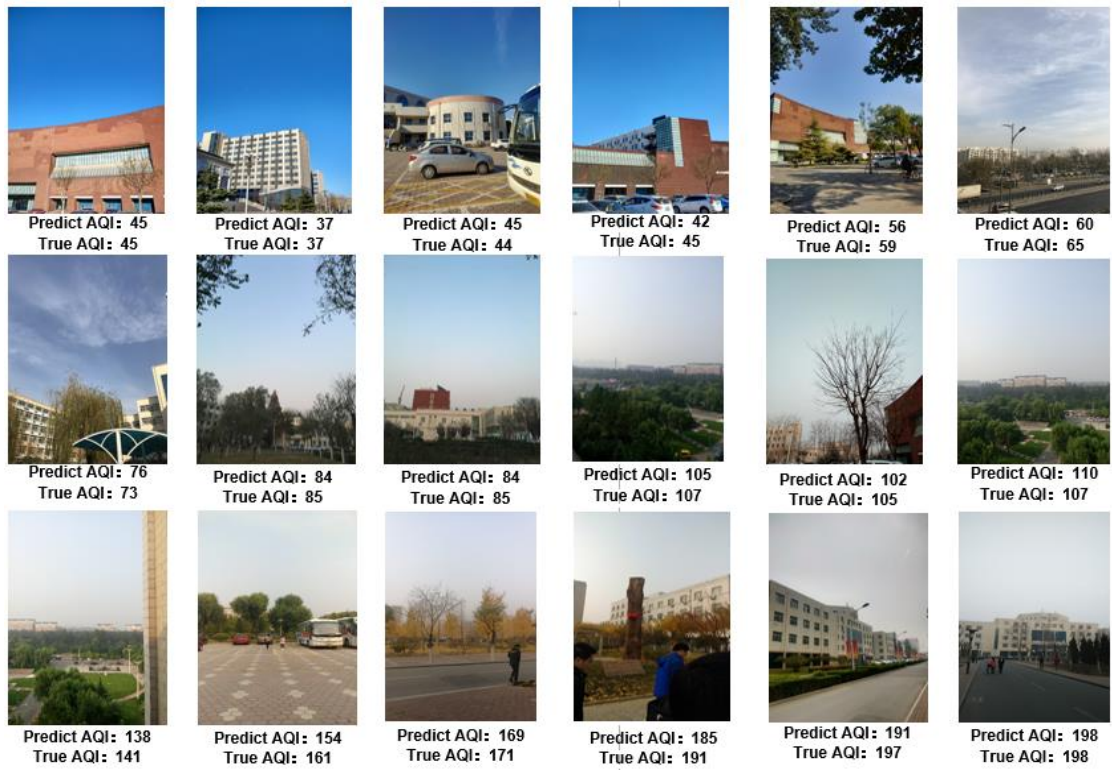

Figure5. Some AQI measurement results of DCSLW-R

Based on the above experiments, we compare with DCSLW-C, DCEW-C, SC-C, DCSLW-R, DCEW-R, SC-R six methods performance of both grade measurement and index measurement, as shown in figure 6. On the whole, the classification method using the double-channel convolutional neural network shows obvious advantages in terms of the accuracy of grade measurement, and the regression method using the double-channel convolutional neural network obtains lower MAE. This advantage is further enhanced with the adoption of self-learning weights.

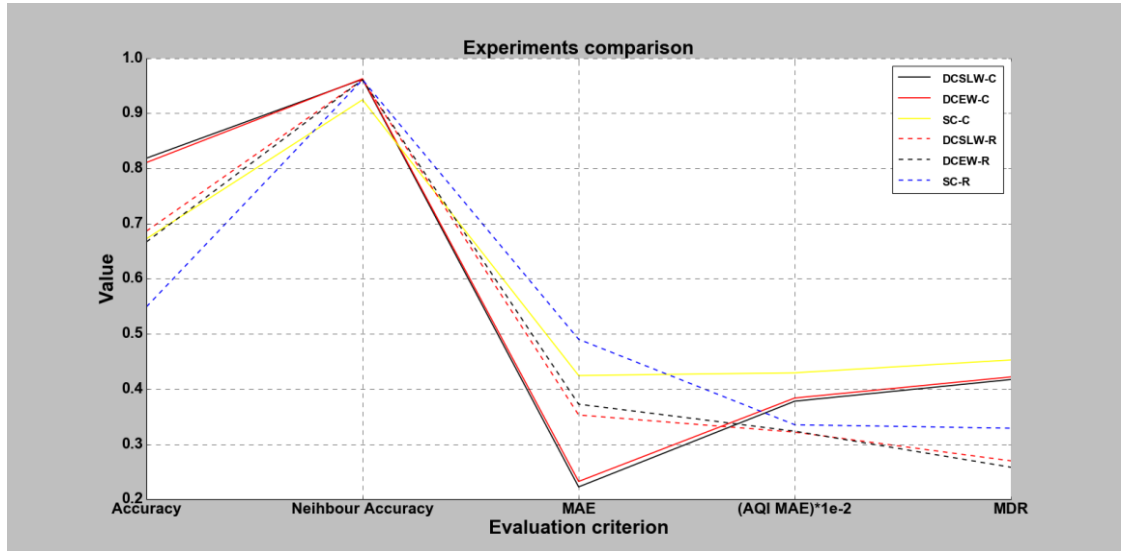

Figure6. Algorithms performance comparison

Meanwhile, in the experiment, we also analyzed the samples that testing failed. DCSLW-C and DCSLW-R partial measurement wrong samples as shown in figure 7. We found that most wrong samples were high air quality grade samples, and the difference between these images with different grades was small, and there was a similarity with the error category. At the same time, due to the label limitations mentioned above, the labels were approximately collected at the nearest time and place monitoring points, so there are certain errors in the label itself, which are also part of the reasons for the measurement errors.

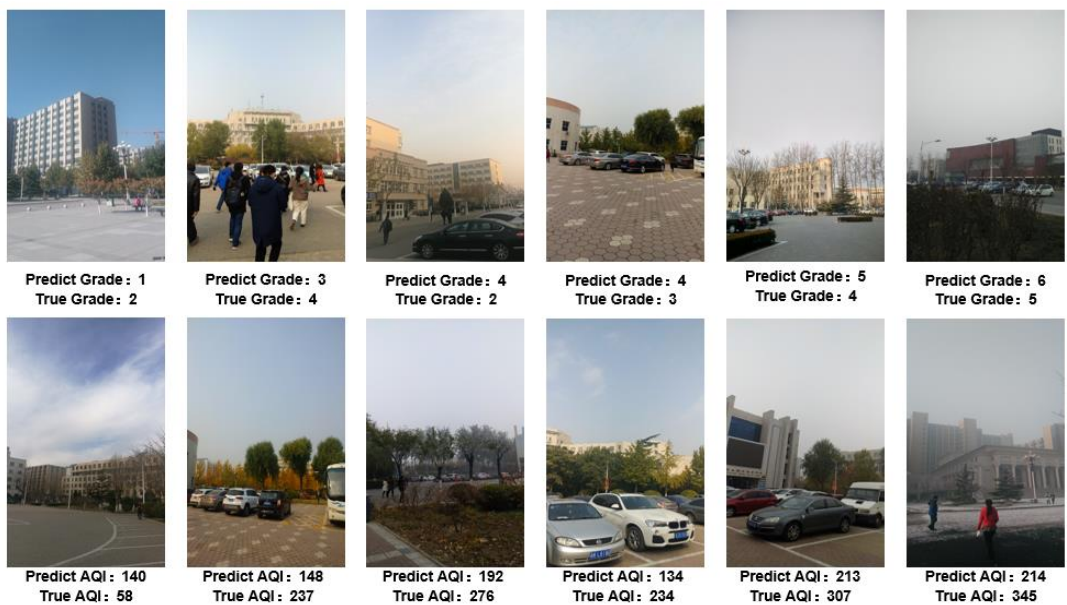

Figure7. Some measurement wrong images

# 5 Conclusion

In this paper, we propose an air quality measurement algorithm based on double-channel convolutional neural network ensemble learning, design a double-channel convolutional neural network structure to measure the air quality grade and index of environmental images. At the same time, we propose a self-learning method of weighted feature fusion. Experimental results show that our method is feasible, achieve certain accuracy and small mean absolute error. On the basis of the double-channel convolutional neural network, the performance is further improved by using the weighted feature fusion method. Compared with other methods, our method achieves better performance. Meanwhile, it is found in the experiment that it is difficult to distinguish the images of adjacent grades and the images of different grades with similar contents. How to recognize such samples will be an important direction of future work.

# References

[1] Zheng Zhang, Huadong Ma, Huiyuan Fu, Xinpeng Wang, Outdoor Air Quality Inference from Single Image, MultiMedia Modeling. 21st International Conference, MMM 2015, Sydney, NSW, Australia, 5-7 Jan. 2015

[2] Chao Zhang, Junchi Yan, Changsheng Li[*], Xiaoguang Rui, Liang Liu, Rongfang Bie, On Estimating Air Pollution from Photos Using Convolutional Neural Network, 24th ACM Multimedia Conference, MM 2016, Amsterdam, United Kingdom, October 15, 2016 - October 19, 2016

[3] Avijoy Chakma, Ben Vizena, Tingting Cao, Jerry Lin, Jing Zhang, Image-based air quality analysis using deep convolutional neural network, 2017 IEEE International Conference on Image Processing (ICIP), Beijing, China, 17-20 Sept. 2017

[4] Nabin Rijal, Gutta Ravi Teja, Tingting Cao, Jerry Lin, Qirong Bo, Jing Zhang, Ensemble of Deep Neural Networks for Estimating Particulate Matter from Images, 3rd IEEE International Conference on Image, Vision and Computing, ICIVC 2018, Chongqing, China, June 27, 2018 - June 29, 2018

[5] Jian Ma, Kun Li, Yahong Han, Jingyu Yang, Image-based Air Pollution Estimation Using Hybrid Convolutional Neural Network, 2018 24th International Conference on Pattern Recognition (ICPR), Beijing, China, 20-24 Aug. 2018

[6] Kaiming He, Jian Sun, Xiaoou Tang, Single Image Haze Removal Using Dark Channel Prior, IEEE Transactions on Pattern Analysis and Machine Intelligence, 2011, 33(12): 2341-2353

[7] Xiaoguang Chen, Yaru Li, Dongyue Li, An Efficient Method for Air Quality Evaluation via ANN-based Image Recognition. 2nd International Conference on Artificial Intelligence and Industrial Engineering (AIIE), Beijing, China, NOV 20-21, 2016

[8] LeCun Y, Boser B, Denker J. S, Henderson D, Backpropagation Applied to Handwritten Zip Code Recognition, Neural Computation, 1989, 1(4):541-551

[9] Lecun Yann, Bottou Ligon, Bengio Yoshua, Haffner Patrick, Gradient-Based Learning Applied to Document Recognition, Proceedings of the IEEE, 1998, 86(11):2278-2323

[10] Geoffrey E.Hinton, Simon Osindero, Yee-Whye The, A fast learning algorithm for deep belief nets. Neural Computation, 2006, 18(7):1527-1554

[11] Alex Krizhevsky, Ilya Sutskever, Geoffrey E.Hinton, Imagenet classification with deep convolutional neural networks, 26th Annual Conference on Neural Information Processing Systems 2012, NIPS 2012, Lake Tahoe, NV, United states, December 3, 2012 - December 6, 2012

[12] Karen Simonyan, Andrew Zisserman, Very Deep Convolutional Networks for Large-Scale Image Recognition, Computer Science, 2014

[13] Christian Szegedy, Wei Liu, Yangqing Jia, Pierre Sermanet, Scott Reed, Dragomir Anguelov, Dumitru Erhan, Vincent Vanhoucke, Andrew Rabinovich, Going deeper with convolutions, IEEE Conference on Computer Vision and Pattern Recognition, CVPR 2015, Boston, MA, United states, June 7, 2015 - June 12, 2015

[14] Kaiming He, Xiangyu Zhang, Shaoqing Ren, Jian Sun, Deep Residual Learning for Image Recognition, 29th IEEE Conference on Computer Vision and Pattern Recognition, CVPR 2016, Las Vegas, NV, United states, June 26, 2016 - July 1, 2016

[15] Nitish Srivastava, Geoffrey E.Hinton , Alex Krizhevsky, Ilya Sutskever, Ruslan Salakhutdinov, Dropout: A Simple Way to Prevent Neural Networks from Overfitting. Journal of Machine Learning Research 2014, 15(1):1929-1958

[16] Ministry of environmental protection of the People's Republic of China. GB3095-2012 Ambient air quality standards. Beijing, 2012.

[17] http://www.bjmemc.com.cn/